\documentclass{amsart}
\usepackage[dvips]{graphicx}
\graphicspath{ {.} }
\setkeys{Gin}{width=\linewidth,keepaspectratio=true}
\usepackage[latin1]{inputenc}
\usepackage{natbib}
\bibpunct{(}{)}{;}{a}{,}{;}
\usepackage[dvipsnames]{xcolor}
\usepackage{xcolor}
\colorlet{RED}{red}
\usepackage{url}
\usepackage{epsfig}
\usepackage[linesnumbered,ruled,vlined]{algorithm2e}
\usepackage{algpseudocode}
\usepackage{rotating}
\usepackage{lscape}
\usepackage{amsmath,amssymb}
\usepackage{numprint}
\npdecimalsign{.} 

\usepackage{mfirstuc}
\usepackage{verbatim}

\newcommand{\Xcomment}[1]{}

\begin{document}

\title[Random-key genetic algorithms]
      {Random-key genetic algorithms:\\ Principles and applications}

\author[M.A. Londe]{Mariana A. Londe}

\address[Mariana A. Londe]{Pontifical Catholic University of Rio de Janeiro,
Rua Marqu\^es de S\~ao Vicente, 225, G\'avea,
Rio de Janeiro, RJ 22453-900, Brazil}

\email[]{mlonde@aluno.puc-rio.br}

\author[L.S. Pessoa]{Luciana S. Pessoa}

\address[Luciana S. Pessoa]{Pontifical Catholic University of Rio de Janeiro,
Rua Marqu\^es de S\~ao Vicente, 225, G\'avea,
Rio de Janeiro, RJ 22453-900, Brazil}

\email[]{lucianapessoa@puc-rio.br}

\author[C.E. Andrade]{Carlos E. Andrade}

\address[Carlos E. Andrade]{AT\&T Labs Research, Bedminster, NJ 07921, USA}

\email[]{cea@research.att.com}

\author[J.F. Gon\c{c}alves]{Jos\'e F. Gon\c{c}alves}

\address[Jos\'e F. Gon\c{c}alves]{Universidade do Porto, Porto, Portugal}

\email[]{jfgoncal@gmail.com}

\author[M.G.C. Resende]{Mauricio G.C. Resende}

\address[Mauricio G.C. Resende]{Center for Discrete Mathematics and 
Theoretical Computer Science, Rutgers University, Piscataway, NJ, 08854 USA}

\email[]{mgcr@berkeley.edu}

\begin{abstract}
A random-key genetic algorithm is an evolutionary metaheuristic for discrete and global optimization. Each solution is encoded as a vector of $n$ random keys, where a random key is a real number randomly generated in the continuous interval $[0, 1)$. A decoder maps each vector of random keys to a solution of the optimization problem being solved and computes its cost.
The benefit of this approach is that all genetic operators and 
transformations can be maintained within the unitary hypercube, 
regardless of the problem being addressed. 
This enhances the productivity and maintainability 
of the core framework.
The algorithm starts with a population of $p$ vectors of random keys. At each iteration, the vectors are partitioned into two sets: a smaller set of high-valued elite solutions and the remaining non-elite solutions. All elite elements are copied, without change, to the next population. A small number of random-key vectors (the mutants) is added to the population of the next iteration. The remaining elements of the population of the next iteration are generated by combining, with the parametrized uniform crossover of \citet{SpeDej91}, pairs of solutions. This chapter reviews random-key genetic algorithms and describes an effective variant called biased random-key genetic algorithms.

\end{abstract}  
\keywords{{Genetic Algorithm}, {Random Keys}, {Optimization}, 
{Random-Key Genetic Algorithm}, {Biased Random-Key Genetic Algorithm}, 
{Metaheuristic}, {Parameter Tuning}.
}  

\date{May 30, 2025.  To appear in \textit{Handbook of Heuristics},
2nd edition, R. Mart\'{\i}, P.M. Pardalos, and M.G.C. Resende, editors,
Springer-Nature, New York, 2025.
}

\maketitle

\section{Genetic algorithm with random keys}
\label{s_rkga}

Genetic Algorithms (GA) are inspired by Darwin's theory of evolution. \citet{Holland1975:genetic_algorithm} introduces this approach, which considers that a possible solution to a combinatorial optimization problem is an individual inside a population of potential solutions. In this population, evolutionary operators are applied at each iteration/generation to obtain new populations and solutions.

A class of GAs is described by \citet{Bean1994:random_keys}, whose solutions are represented by a vector of random keys, called a chromosome. A \textit{random key} is a real number belonging to the interval~$[0,1)$. This makes the evolutionary process of the \textit{random-key genetic algorithms} (RKGA) 
problem-independent,
as the evolutionary operators can be specified for the chromosomes, which allows for a generalization of the framework. The quality of a solution is calculated by the \textit{decoder} procedure, which maps a chromosome to a fitness value. In \citep{Bean1994:random_keys}, the decoding procedure simply orders the elements of the vector of random keys, which produces a permutation of the indices of the sorted elements.

In the RKGA evolutionary process, the algorithm starts with an initial population $\mathcal{P}$ of vectors of $n$ random keys. In each generation, the vectors are partitioned into a set of the best solutions $\mathcal{P^e} < |\mathcal{P}|/2$, called the elite set, and a set with the remainder of non-elite solutions of the population. All elite solutions are copied, unchanged, into the population of the next generation. This elitism characterizes the Darwinian principle in the RKGA. Next, a set $\mathcal{P^m}$ of randomly created vectors is introduced into the next population. This set is comprised of \textit{mutants}, which play the same role as the mutation operators of classical GAs, i.e., they help increase populational diversity and avoid convergence to non-global local optimum. To complete the elements of the next population, $|\mathcal{P}| - |\mathcal{P^e}| - |\mathcal{P^m}|$ are generated by combining solutions from the entire population using a parametrized uniform crossover \citep{Spears1991:multi_point_crossover}. In this crossover, consider $a$ and $b$ to be the vectors chosen for mating, and let $c$ be the offspring produced. Random key $c[i]$, the $i$-th component of the offspring vector, receives the $i$-th key of one of its parents. It receives the key $a[i]$ with probability $\rho_a$ and $b[i]$ with probability $\rho_b = 1 - \rho_a$.

\section{Biased random-key genetic algorithms}
\label{s_brkga}

A \textit{biased random-key genetic algorithm} (BRKGA), meanwhile, also uses an elitist strategy in the mating procedure, not only regarding the reproduction of the elite set. 

\begin{figure}
\centering
\includegraphics[scale = 0.5]{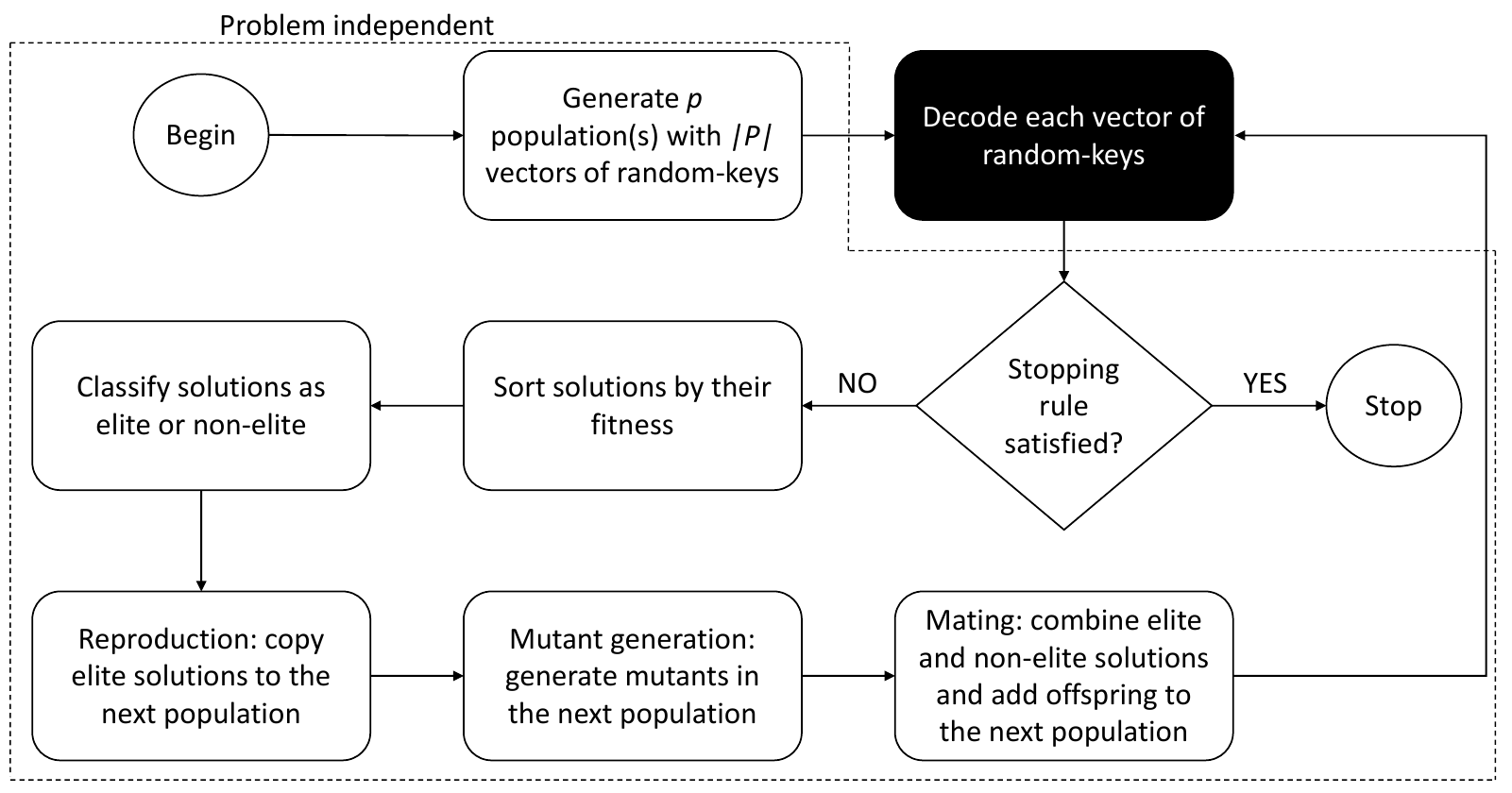}
\caption{Flowchart detailing BRKGA process. Adapted from \citet{Goncalves2011:BRKGA}.}
\label{fig:flowchart_brkga}     
\end{figure}

In BRKGA, one parent is always chosen from the elite set. When performing the coin toss to choose whose parent passes the gene to the offspring, the gene of the elite parent has a probability of being inherited $\rho > 0.5$. This means that the offspring has a higher chance of inheriting the keys of the elite parent, and of retaining substructures of a good solution while still allowing the insertion of substructures of a not-so-good solution. This difference between the two algorithms tends to result in BRKGA outperforming Bean's algorithm \citep{Goncalves2014:comparison_biased_unbiased_RKGA}. As a result, most of the literature published after the introduction of BRKGA tends to use the biased variant. A flowchart detailing the BRKGA framework can be seen in Figure~\ref{fig:flowchart_brkga}. In it, the problem-independent structure of classical BRKGA is highlighted. The decoding procedure in black is problem-dependent, as it makes the correlation between a random-key vector and a feasible problem solution.

The double elitism of BRKGA leads to rapid convergence to high-quality solutions. Alongside its problem-agnostic approach, it has led to a significant increase in use, with over~250 papers published until~2025 \citep{Londe2025:BRKGA_review,Londe2025:BRKGA_early_years}.
The BRKGA was formally introduced in \citet{Goncalves2011:BRKGA}, but some of its characteristics can be seen in the early 2000s \citep{Buriol2010:road_congestion,Ericsson2002:Genetic_alg_OSPF,Goncalves2002:hybrid_assembly_line_balancing}. Since then, several applications, hybridizations, and features have been described in the literature. Among the problems in which BRKGA was applied since 2020, we have:

\begin{itemize}
    \item \textit{Scheduling:} 
    \citet{Abreu2021:scheduling_routing_makespan}, 
    \citet{Abreu2022:customer_order_scheduling}, 
    \citet{Bolsi2022:heuristic_workforce_allocation_scheduling},  
    \citet{Fontes2023:bi_brkga_schedule_quay_cranes}, 
    \citet{Fontes2023:job_shop_transp_resources},     
    \citet{Huang2024:coevo_hybrid_flowshop},
    \citet{Ibarra-Rojas2025:timetabling_scheduling}, 
    \citet{Kong2020:parallel_scheduling_deterioration_maintenance},
    \citet{Kong2020:scheduling_uncertain_rolling_deterioration}, 
    \citet{Kummer2022:home_health_care}, 
    \citet{Liu2024:multipop_coevo}, 
    \citet{Maecker2023:unrelated_parallel_machine_scheduling}, 
    \citet{Queiroga2021:ils_scheduling}, 
    \citet{Rocholl2021:scheduling_parallel_common_due_date}, 
    \citet{Silva-Soto2021:timetabling_bus_lines}, 
    \citet{Soares2022:resource_constrained_scheduling_setup}, 
    \citet{Soares2020:scheduling_identical_parallel}, 
    \citet{Xie2022:adaptive_brkga_cloud_workflow_scheduling}, 
    \citet{Yang2025:knowledge_brkga}, 
    \citet{Yu2023:supply_scheduling}, 
    \citet{Zhang2022:parallel_batch_scheduling_packing}, and   
    \citet{Zhao2023:distributed_sequential_assembly_process};
    \item \textit{Network Configuration:}    
    \citet{Mikulski2021:energy_storage}, 
    \citet{Pinto2020:routing_assignment}, 
    \citet{Raposo2020:network_reconfiguration_energy_loss}, and
    \citet{Usberti2024:fault_indicators};
    \item \textit{Location:} 
    \citet{Freitas2023:brkga_two_lvl_hub},  
    \citet{Johnson2020:cover_by_pairs}, 
    \citet{Londe2021:pnext_center}, 
    \citet{Mauri2021:multiproduct_facility_location}, 
    \citet{Mendoza-Andrade2025:capacitated_location_tree}, and 
    \citet{Villicana2022:mobile_labs_covid19_testing};
    \item \textit{Cutting and packing:} 
    \citet{AmaroJunior2021:minimum_time_cut_path},    
    \citet{Goncalves2020:two_dimensional_cutting_defects},  
    \citet{Lu2023:additive_manufacturing_production}, 
    \citet{Mandal2024:2d_strip_packing},     
    \citet{Oliveira2022:diversity_scenario_gen},
    \citet{Oliveira2022:two_dimensional_non_guillotine_cutting}, 
    \citet{SouzaQueiroz2020:irregular_knapsack}, and
    \citet{Zhou2024:airfreight_shipment};
    \item \textit{Traveling Salesman and Vehicle Routing:} 
    \citet{Junior2023:a_brkga_laser_cutting}, 
    \citet{Carrabs2021:set_orienteering}, 
    \citet{Chagas2020:traveling_salesman_loading}, 
    \citet{Chagas2021:bi_objective_traveling_thief}, 
    \citet{Chaves2024:adaptive_ftsp}, 
    \citet{Damm2025:adaptive_multiobj_brkga}, 
    \citet{Huang2020:bi_objective_road_disruptions}, 
    \citet{Huang2020:model_urban_road_network_disruptions}, 
    \citet{Ibarra-Rojas2021:routing_equity}, 
    \citet{Marques2023:two_phase_multi_obj_UAV_routing},
    \citet{Oliveira2024:public_transport}, 
    \citet{Santiyuda2024:biobjective_traveling_thief}, 
    \citet{Schenekemberg2022:dial_a_ride}, 
    \citet{Schenekemberg2024:dial_a_ride_hybrid}, and
    \citet{Schuetz2022:robot_traject_planning};
    \item \textit{Clustering:} 
    \citet{Brito2020:stratified_sampling}, 
    \citet{Fadel2021:statistical_disclosure_control},
    \citet{Martarelli2020:feature_selection}, and
    \citet{Xu2022:ABKRGA_local_search};
    \item \textit{Graph problems:} 
    \citet{Lima2022:matheuristic_broadcast_time}, 
    \citet{Londe2022:root_sequence_index}, 
    \citet{Londe2025:multistage_rsi_ejor}, 
    \citet{Melo2023:brkga_quasi_clique_partitioning}, 
    \citet{Pinto2021:maximum_quasi_clique_local_search}, 
    \citet{Silva2024:grundy_chromatic_number}, and
    \citet{Silva2023:brkga_chordal_completion};
    \item \textit{Automatic tuning of parameters in heuristics:} 
    \citet{ErzurumCicek2021:artificial_network_time_series}, 
    \citet{Falls2022:ocean_model_parameter_optimization}, 
    \citet{Japa2023:hybrid_brkga_hyperparameter_neural_nets}, and
    \citet{Sun2022:bayesian_network_structure};
    \item \textit{Other applications:} 
    \citet{Morgan2023:multispacecraft_maneuvers}, 
    \citet{Ochoa2021:search_trajectory_networks}, 
    \citet{Pan2021:covid19_spread}, 
    \citet{Pastore2022:bezier_brkga_topology}, 
    \citet{Pinacho-Davidson2020:minimum_capacitated_dominationg_set}, 
    \citet{Ramos2025:multiobjective_residential_energy}, 
    \citet{Reixach2025:longest_common_square}, and
    \citet{Silva2023:BRKGA_maximum_diversity}.
\end{itemize}

\section{A model for the implementation of a BRKGA}
\label{s_modelo}

Algorithm~\ref{a_brkga} shows a pseudo-code of a BRKGA for the minimization of $f(x)$, where $x \in X$ and $X$ is a discrete set of solutions and $f: X \rightarrow \mathbb{R}$. This implementation is a multi-start variant of a BRKGA where several populations are evolved in sequence, and the best solution among all in the population is returned as the output of the algorithm.

\linespread{1.0}
\begin{algorithm}[t]
   \SetAlgoLined
\texttt{BRKGA}($|\mathcal{P}|,|\mathcal{P}_e|,|\mathcal{P}_m|,n, \rho_a$)\\
   Initialize value of the best solution found: $f^*  \gets \infty$\;
   \While{stopping criterion not satisfied}{
   Generate a population $\mathcal{P}$ with $|\mathcal{P}|$ vectors of $n$ random keys\;
   \While{restart criterion not satisfied}{
	Evaluate the cost of each new solution in $\mathcal{P}$\;
	Partition $\mathcal{P}$ into two sets: $\mathcal{P}_e$ and $\mathcal{P}_{\bar{e}}$\;
        Initialize population of next generation: $\mathcal{P}^+ \gets \mathcal{P}_e$\;
        Generate set $\mathcal{P}_m$ of mutants, each mutant with
        $n$ random keys\;

        Add $\mathcal{P}_m$ to population of next generation: $\mathcal{P}^+ \gets \mathcal{P}^+ \cup \mathcal{P}_m$\;
        \ForEach{$i \gets 1$ \KwTo $|\mathcal{P}|-|\mathcal{P}_e|-|\mathcal{P}_m|$}{
        	Select parent $a$ at random from $\mathcal{P}_e$\;
        	Select parent $b$ at random from $\mathcal{P}_{\bar{e}}$\;
                \ForEach{$j \gets 1$ \KwTo $n$}{
                      Throw a biased coin with probability $\rho > 0.5$ of resulting heads\;
                      \lIf{heads}{
			    $c[j] \gets a[j]$
			}
			\lElse{$c[j] \gets b[j]$}
                }
                Add offspring $c$ to population of next generation: $\mathcal{P}^+ \gets \mathcal{P}^+ \cup \{ c \}$\;
	}
        Update population: $\mathcal{P} \gets \mathcal{P}^+$\;
   Find best solution $\chi^+$ in $\mathcal{P}$: $\chi^+ \gets \mathbf{argmin} \{ f(\chi) \;|\; \chi \in \mathcal{P} \}$\;
   \If{$f(\chi^+) < f^*$}{
	$\chi^* \gets \chi^+$\;
	$f^* \gets f(\chi^+)$\;
   }
   }
}
   \Return{$\chi^*$}
   \caption{Model for biased random-key genetic algorithm with restart.}
   \label{a_brkga}
\end{algorithm}
\linespread{1.5}

In line~2, the value $f^*$ of the best solution found is initialized to a large value, i.e., not smaller than $f(x^0)$, where $x^0 \in X$ is some feasible solution to the problem. Evolution of each population is done in lines~3 to~28. The algorithm halts when a stopping criterion in line~3 is satisfied. This criterion can be, for example, the number of evolved populations, the total time, or the quality of the best solution found.

In line~4, the population being evolved is initialized with $|\mathcal{P}|$ vectors of random keys. Evolution of population $\mathcal{P}$ takes place in lines~5 to~27. This evolution ends when a restart criterion is satisfied in line~5. This criterion can be, for example, a maximum number of generations without improvement in the value of the best solution in $\mathcal{P}$. At each generation, or iteration, the following operations are carried out: In line~6, all new solutions (offspring and mutants) are decoded and their costs evaluated. Note that each decoding and evaluation in this step can be computed simultaneously, i.e., in parallel. In line~7, population $\mathcal{P}$ is partitioned into two subpopulations $\mathcal{P}_e$ (elite) and $\mathcal{P}_{\bar e}$ (non-elite), where $\mathcal{P}_e$ is such that $|\mathcal{P}_e| < |\mathcal{P}|/2$ and contains $|\mathcal{P}_e|$ of the best solutions in $\mathcal{P}$ and $\mathcal{P}_{\bar e}$ consists of the remaining solutions in $\mathcal{P}$, that is $\mathcal{P}_{\bar e} = \mathcal{P} \setminus \mathcal{P}_{e}$. $\mathcal{P}^+$ is the population of the next generation. It is initialized in line~8 with the elite solutions of the current generation. In line~9, the mutant subpopulation $\mathcal{P}_{m}$ is generated. Each mutant is a vector of $n$ random keys. The number of generated mutants in general is such that $|\mathcal{P}_m| < |\mathcal{P}|/2$. This subpopulation is added to the population $\mathcal{P}^{+}$ of the next generation in line~10.

With $|\mathcal{P}_e| + |\mathcal{P}_m|$ vectors inserted in population $\mathcal{P}^+$, it is necessary to generate $|\mathcal{P}| - |\mathcal{P}_2| - |\mathcal{P}_m|$ new offspring to complete the $|\mathcal{P}|$ vectors that form population $\mathcal{P}^{+}$. This is done in lines~11 to~20. In lines~12 and~13 parents $a$ and $b$ are chosen, respectively, at random from subpopulations $\mathcal{P}_e$ and $\mathcal{P}_{\bar e}$. The generation of offspring $c$ from parents $a$ and $b$ occurs in lines~14 to~18. A biased coin (with probability $\rho > 0.5$ of flipping to heads) is thrown $n$ times. If the $i$-th toss is a head, the offspring inherits the $i$-th key of parent~$a$. Otherwise, it inherits the $i$-th key of parent~$b$. After the offspring is generated, $c$ is added to population $\mathcal{P}^{+}$ in line~19.

The generation of $\mathcal{P}^{+}$ ends when it consists of $|\mathcal{P}|$ elements. In line~21, $\mathcal{P}^{+}$ is copied to $\mathcal{P}$ to start a new generation. The best solution in the current population in evolution is computed in line~22, and if its value is better than all solutions examined so far, the solution and its cost are saved in lines~24 and~25 as $\chi^*$ and $f^*$, respectively. $\chi^*$, the best solution found across all populations, is returned by the algorithm in line~29.

\section{Optional procedures of the framework}
\label{s_features}

There are several ways to modify the BRKGA algorithm in order to increase solution quality, convergence, diversity, and other criteria. One of these is the use of the restart. In BRKGA, the restart or reset of the population is the elimination of all current individuals and the re-creation of the entire population with random chromosomes. This is done, generally, after a $\text{\slshape I}_{ipr}$ number of iterations without improvement in the best overall solution $\chi*$, similar to the restart strategy for
GRASP with path-relinking \citep{ResRib11a}, has been shown to increase individual diversity and prevent premature convergence \citep{Pandey2014:premature_convergence,Toso2015:C++_app_BRKGA}. 

A partial restart is also possible. The shaking method \citep{Andrade2019:flowshop_scheduling} is akin to a perturbation of the Iterated Local Search (ILS) algorithm \citep{Lourencco2003:ils}, and acts by modifying a part of the genes of the elite solutions while the non-elite chromosomes are completely restarted. That means some of the structures gained in the evolutionary process are preserved in the elite set, while the complete population has guaranteed diversity. The shake operator is usually applied after $\text{\slshape I}_s$ iterations without improvement in the best overall solution $\chi*$, and, like the ILS perturbation, has strength parameters $\text{\slshape SK}_l$, indicating the lower bound, and $\text{\slshape SK}_u$, corresponding to the upper bound.

Another common feature is the island model \citep{Whitley1999:island_model}. Inspired by divergent evolution, the island model is the simultaneous evolution of $p$ parallel populations. The idea is that parallel evolution will lead to different characteristics in the solutions, thereby enabling a more extensive exploration of the solution space. Every $g$ generations, $i$ elite individuals are exchanged among populations, which increases diversity and helps spread novel and interesting structures among the solutions. 

Many authors also attempt to introduce interesting structures into the initial population by injecting warm start solutions. This addition is shown to increase the performance and convergence of the method. In general, one or more good solutions are created with other heuristics \citep{Abreu2021:scheduling_routing_makespan,Junior2023:a_brkga_laser_cutting,Chagas2021:bi_objective_traveling_thief,Londe2022:root_sequence_index,Martarelli2020:feature_selection,Silva2023:BRKGA_maximum_diversity,Xie2022:adaptive_brkga_cloud_workflow_scheduling} or mathematical models \citep{Chagas2020:traveling_salesman_loading,Schuetz2022:robot_traject_planning} and then inserted into the initial population. In some cases, the entire initial population can be created using alternative methods, as is done in \citep{Zhang2022:parallel_batch_scheduling_packing}, where the solutions have specific characteristics that are not easily obtained in randomly generated individuals.

One more recent addition to the BRKGA framework is the use of multi-parent crossover \citep{Andrade2021:BRKGA_MP_IPR}. In it, several parents are considered in the mating process, and the choice of inheritance is done using a ranking function. Consider the following example: we have three total parents, denoted as $a$, $b$, and $c$. Of those, both $a$ and $b$ belong to the elite set, while $c$ belongs to the non-elite set. For multi-parent crossover, we would rank those parents according to their fitness values, so that the first in order would have better fitness than the following ones, and so forth. In this case, considering a minimization problem, we observe that $f(a) < f(b) < f(c)$. The ranking of solutions would, thus, be $r(a) = 1$, $r(b) = 2$, and $r(c) = 3$. With these values, the procedure uses a bias function $\Phi(r)$ \citep{Bresina1996:stochastic_sampling} to calculate the probabilities of each parent being chosen in the mating. The bias function is non-increasing, so parents with better fitness have a higher chance of being chosen and can belong to several classes, such as exponential, linear, logarithmic, and others. 

One last optional procedure is Implicit Path-Relinking \citep{Andrade2021:BRKGA_MP_IPR}. The classical path-linking is an intensification strategy that explores the neighborhood in the path between two distinct solutions \citep{Ribeiro2012:PR,Glover1997:PR}. Typically, path-relinking is problem-dependent, due to the exploration of the intermediate solutions. However, the implicit variant leverages the random-key structure of BRKGA to perform the procedure within the random-key solution space, with differences depending on the type $typ$ of the decoding process. This procedure works by first selecting two solutions from different populations using method $sel$, which can be the best or randomly chosen among the elite individuals. The selected individuals must have a minimum distance of $md$, which is calculated by either the Hamming (direct IPR) or Kendall Tau (permutation IPR) distances. A percentage $cp\%$ of the total pairs of elite chromosomes are tested as candidates from IPR, and if no pairs have the minimum distance, then the IPR procedure is not executed. This limitation of candidate pair testing is done to decrease the computational burden of the algorithm. Two other parameters are used in IPR. Parameter $ps\%$ is the size of the path between the two candidates that is evaluated at each IPR iteration, given as a percentage of the chromosome size. In each test, a block of genes of size $bs$ is modified in the base solution to become similar to the one in the guide solution, thus obtaining intermediate solutions that are then explored.

\section{Hybridizing a random-key genetic algorithm}
\label{s_hybrid}

To increase performance and/or adapt the algorithm to problems with specific characteristics, the authors often incorporate additional methods or frameworks to work in conjunction with BRKGA. This practice is very common for BRKGA, especially when the addition is a local search.

The use of local search (LS) in a genetic framework is viewed as an application of Lamarck's theory of evolution, in which individuals pass on the characteristics they acquire during their lifetime to their descendants \citep{Cotta2018:memetic_alg}. It is particularly common for BRKGA, as it has been shown to significantly increase solution quality and convergence \citep{Londe2025:BRKGA_review}. The use of local search can be within \citep{Londe2022:root_sequence_index,Maecker2023:unrelated_parallel_machine_scheduling} or outside the decoding process, with the latter case applied to all chromosomes in every iteration \citep{Brito2022:k_medoids_clustering,Silva2023:BRKGA_maximum_diversity}, only the best incumbent solution after evolution is complete \citep{Chagas2021:bi_objective_traveling_thief,Mauri2021:multiproduct_facility_location}, or to special chromosomes in some generations \citep{Abreu2022:customer_order_scheduling,Carrabs2021:set_orienteering,Japa2023:hybrid_brkga_hyperparameter_neural_nets,Londe2021:pnext_center}. Established methods can also be used as LS, such as Variable Neighborhood Search  \citep{Chaves2024:adaptive_ftsp,Lu2023:additive_manufacturing_production,Schenekemberg2022:dial_a_ride,Soares2022:resource_constrained_scheduling_setup,Soares2020:scheduling_identical_parallel}, Clustering Search \citep{Chaves2016:hybrid_tool_switches}, ILS \citep{Silva2019:multicommodity_tsp}, IGA \citep{Abreu2021:scheduling_routing_makespan}, and INLS \citep{Xu2022:ABKRGA_local_search}.

When not performed as a local search, hybridization is typically used to adapt the framework to specific problem characteristics. Some examples are:

\begin{itemize}
    \item \textit{Multi-objective optimization:}
    \citet{Bierbusse2025:aircraft_manufaturing}, 
    \citet{Bolsi2022:heuristic_workforce_allocation_scheduling}, 
    \citet{Chagas2021:bi_objective_traveling_thief}, 
    \citet{Damm2025:adaptive_multiobj_brkga}, 
    \citet{Fontes2023:bi_brkga_schedule_quay_cranes}, 
    \citet{Fontes2023:job_shop_transp_resources}, 
    \citet{Ibarra-Rojas2025:timetabling_scheduling}, 
    \citet{Liu2024:multipop_coevo}, 
    \citet{Marques2023:two_phase_multi_obj_UAV_routing}, 
    \citet{Mikulski2021:energy_storage}, 
    \citet{Ramos2025:multiobjective_residential_energy}, 
    \citet{Raposo2020:network_reconfiguration_energy_loss}, 
    \citet{Santiyuda2024:biobjective_traveling_thief}, and
    \citet{Silva-Soto2021:timetabling_bus_lines};
    \item \textit{Matheuristics:} 
    \citet{Lima2022:matheuristic_broadcast_time}, 
    \citet{Pinacho-Davidson2020:minimum_capacitated_dominationg_set}, 
    \citet{Pinto2021:maximum_quasi_clique_local_search}, and
    \citet{Sun2022:bayesian_network_structure};
    \item \textit{Machine learning:} 
    \citet{Junior2023:a_brkga_laser_cutting}, 
    \citet{Damm2025:adaptive_multiobj_brkga}, 
    \citet{Pan2021:covid19_spread}, 
    \citet{Schenekemberg2022:dial_a_ride}, 
    \citet{Schenekemberg2024:dial_a_ride_hybrid}, 
    \citet{Xie2022:adaptive_brkga_cloud_workflow_scheduling}, 
    \citet{Xu2022:ABKRGA_local_search}, and
    \citet{Yang2025:knowledge_brkga};
    \item \textit{Stochastic optimization and scenario generation:} 
    \citet{Huang2024:coevo_hybrid_flowshop}, 
    \citet{Londe2021:pnext_center}, 
    \citet{Oliveira2019:matheuristic_car_rental_stochastic}, 
    \citet{Oliveira2021:scenario_generation}, and
    \citet{Oliveira2022:diversity_scenario_gen};
\end{itemize}

\section{Customizing a RKGA}
\label{s_spec}

To customize an RKGA, one needs to define how a solution is represented, or encoded, and how the corresponding decoding procedure is done. Also, it is necessary to choose and customize any hybrid methods, if applicable. After this, the algorithm's parameters must be tuned.

Since each solution is represented as a vector of $n$ random keys, it is necessary only to specify a value for $n$ for the encoding. It may represent the number of vertices in a graph, or the total number of jobs and pauses in parallel machines, or the possible locations for facility installation, or the number of parameters to be analyzed, for example.

The decoder is a deterministic procedure that takes as input the chromosome of $n$ random keys and produces as output the solution of the problem alongside the corresponding fitness value. We can define three types of decoding procedures: indicator, permutation, and a mix of both. In a permutation decoder, the solution is obtained by sorting the random keys and constructing a feasible solution with the permutation of the indices. Examples of problems that often have permutation-based decoders are Location \citep{Londe2021:pnext_center,Mauri2021:multiproduct_facility_location}, Cutting and Packing \citep{AmaroJunior2021:minimum_time_cut_path,Oliveira2022:two_dimensional_non_guillotine_cutting}, Vehicle Routing \citep{Ibarra-Rojas2021:routing_equity,Schenekemberg2022:dial_a_ride}, Traveling Salesman Problem \citep{Junior2023:a_brkga_laser_cutting,Chagas2021:bi_objective_traveling_thief}, and Graph Problems \citep{Lima2022:matheuristic_broadcast_time,Silva2023:brkga_chordal_completion}.

An indicator decoder, in contrast, obtains meaning directly from the values of the random keys. For example, consider that one wishes to assign each point, represented by one random key, to one of four classes. In this case, the algorithm may consider that the interval $[0,1)$ is divided into four equal intervals of size $0.25$, with the first class being indicated by values in the interval $[0,0.25)$, the second class by the interval $[0.25,0.50)$, and so forth. For solution $c$, if the $i$-th gene has value $0.25 \leq c[i] < 0.50$, then the corresponding $i$-th element belongs to the second class. The use of indicator decoders is common in Network Configuration \citep{ Andrade2021:BRKGA_MP_IPR,Andrade2022:PCI_MO, Pedrola2013:hybridizations_regenerator_placement_dimensioning,Raposo2020:network_reconfiguration_energy_loss}, Clustering \citep{Martarelli2020:feature_selection,Oliveira2017:hybrid_constrained_clustering}, and Parameter Optimization \citep{ErzurumCicek2021:artificial_network_time_series, Falls2022:ocean_model_parameter_optimization, Japa2023:hybrid_brkga_hyperparameter_neural_nets}.

Sometimes, a problem has characteristics that make a pure indicator- or permu\-ta\-tion-based representation inadequate. When this happens, the authors tend to mix those two, generally by partitioning the chromosome into two or more parts. One example of this practice is done in \citep{Villicana2022:mobile_labs_covid19_testing} to locate mobile labs, in which the first genes are sorted to obtain the order of service points, while the last genes indicate an increase of the area of service radius. Another example is \citep{Freitas2023:brkga_two_lvl_hub} for a two-level hub location problem, where the random-key interval is divided by the number of desired hubs. For the genes in each interval, the values are sorted, with the order of the directed graph being indicated by the ordered genes and the first gene corresponding to the hub.
In \citep{Andrade2015:wireless_backhaul}, a complex problem related to wireless network design is addressed using a five-part chromosome. The first two sections consist of random keys that indicate the pertaining and parameterization of the solution. The next two sections are used to establish the evaluation order of the deployed equipment and its relation. Lastly, the final section serves as indicators to determine the depths of the network trees.

\section{Tuning an RKGA}
\label{s_tuning}

Table~\ref{tab:params_values} presents all the parameters that were commented on in the traditional BRKGA and the optional procedures, together with their recommended values from the current literature. One can note that, if all procedures are desired, then almost~$20$ parameters need to be specified to run a BRKGA. 

\begin{sidewaystable}[ph]
\vspace*{400pt}
\caption{BRKGA parameters and recommended ranges}
\label{tab:params_values} 
\begin{tabular}{lr}
\textbf{Parameter} & \textbf{Recommended value} \\
\noalign{\smallskip}\hline\noalign{\smallskip}
$|\mathcal{P}|$: size of population & $\min \{ [100, 500], \lfloor 10 \times n \rfloor \}$ \\
$\mathcal{P_e}\%$: percentage of elite partition of population & $[0.1, 0.5]$ \\
$\mathcal{P_m}\%$: percentage of mutant partition of population & $[0.1, 0.5]$\\
$\rho$: probability of inheriting key from elite parent & $[0.5,1]$\\
$p$: number of parallel populations & $\{ 1, \ldots, 3 \}$\\
$g$: frequency for population interchange &  $[50, 500]$\\
$i$: number of elite migrants between parallel populations & $\{1,2\}$\\
$\text{\slshape I}_r$: iterations without improvement for restart & $[200,500]$\\
$\text{\slshape I}_s$: iterations without improvement for shaking & $[20,100]$\\
$\text{\slshape SK}_l$: lower bound for shake intensity & $[0.1, 0.5]$\\
$\text{\slshape SK}_u$: upper bound for shake intensity & $[0.5, 0.9]$\\
$\pi_t$: total number of parents in multi-parent crossover & $[3, 10]$\\
$\pi_e$: total number of elite parents in multi-parent crossover & $[1, 7],  \pi_e < \pi_t$\\
$\Phi(r)$: type of bias function for multi-parent crossover & $\{1 / \log (r + 1), r^{-1}, r^{-2}, r^{-3}, e^{-r}, 1/\pi_t \}$\\
$sel$: individual selection method for path-relinking & $[BESTSOLUTION, RANDOMELITE]$\\
$cp\%$: percentage of chromosome pairs to be tested for path-relinking & $[50, 500]$\\
$md$: minimum distance between chromosome pairs for path-relinking & $[0.0, 0.30]$\\
$bs$: block size for path-relinking & $1.00$\\
$ps\%$: path size for path-relinking & $[0.01, 1]$\\
$\text{\slshape I}_{ipr}$: frequency of path-relinking & $[50, 500]$\\
\noalign{\smallskip}\hline
\end{tabular}
\end{sidewaystable}

Due to the number of parameters, and how much they affect performance, authors have used many strategies for parameter configuration, either before or during the execution of experiments. When done before, this is called \textit{offline tuning}, and is widely used for metaheuristics \citep{Eiber1999:parameter_control}.

For BRKGA, authors have used several offline strategies. One of the simplest is a grid search, also known as the full factorial design of experiments. Grid search evaluates all possible combinations of given discrete parameters. However, the complexity of the method increases with each parameter and each candidate range, making it costly to perform if there are many parameters with their own interactions and co-dependence~\citep{Feurer2019:hyperparameter}. Some studies that used grid search are \citep{Carrabs2021:set_orienteering,Damm2016:field_tech_scheduling,Pessoa2017:tree_hubs_location,Rochman2017:BRKGA_GD_CVRPTW}.

Other authors utilize parameters suggested by similar groups of problem instances in the literature. However, the performance of a particular algorithm with a specific configuration on some instances is of limited utility, and the generalization of configurations should be done cautiously \citep{Karimi2022:machine_learning}. The use of literature suggestions is done in \citep{ErzurumCicek2021:artificial_network_time_series,Oliveira2021:scenario_generation,Zudio2018:BRKGA_VND_bin_packing}.

Lastly, one may use an automated method of parameterization. For BRKGA, authors seem to prefer Iterated F-Race \citep{LopezIbanez2016:irace}, used recently in papers such as \citep{Andrade2015:wireless_backhaul,Andrade2019:flowshop_scheduling,Andrade2021:BRKGA_MP_IPR,Andrade2022:PCI_MO,Londe2021:pnext_center,Londe2022:root_sequence_index,Londe2025:multistage_rsi_ejor,Mauri2021:multiproduct_facility_location}.

The parameters may also be tuned during the execution of the algorithm. This process, known as \textit{online parameter control}, falls within the class of machine learning techniques and has been applied to BRKGA in two distinct approaches. In both cases, the parameters are updated at each algorithm generation, aiming to balance intensification and exploitation during evolution. In \citep{Junior2023:a_brkga_laser_cutting,Chaves2018:capacitated_centered_clustering_local_search,Silva2019:multicommodity_tsp,Xu2022:ABKRGA_local_search}, the authors use the Adaptive BRKGA (A-BRKGA), which updates parameters for population size, elite proportion, mutant proportion, crossover probability, and maximum number of generations at each iteration. It also utilizes two self-adaptive parameters, denoted as $\alpha$ and $\beta$, where $\alpha$ indicates the number of members from the elite restricted list that evolve, and $\beta$ represents the probability of perturbing individuals that are too similar. At each generation, the values of population size, mutant proportion, and $\alpha$ decrease, while elite proportion increases. $\beta$ and crossover probability are incorporated as genes into the chromosomes, and are thus always modified randomly. 

The other online approach is called BRKGA-QL \citep{Chaves2024:adaptive_ftsp,Schenekemberg2022:dial_a_ride}. It uses a Q-Learning algorithm \citep{Watkins1992:q_learning} to control the same parameters of A-BRKGA in an environment of actions, states, and rewards. Simply put, each possible parameter modification is mapped to a possible reward, and the algorithm chooses one of the actions, with a $\eta$ probability of selecting the action with the highest reward in terms of intensification. The value of $\eta$ decays during evolution, so that in later generations, actions that increase diversification tend to be chosen. BRKGA-QL was extended to work with multi-objective problems \citep{Damm2025:adaptive_multiobj_brkga}, and had a new reward scheme and diversity methods introduced in \citep{Schenekemberg2024:dial_a_ride_hybrid}.

In \citep{Zambelli2022:brkga_tuning}, the author also studies online approaches for BRKGA. It is demonstrated that a random parameter strategy can yield better results compared to more elaborate schemes. This strategy defines upper and lower bounds for parameters $\mathcal{P}$, $\mathcal{P^e}$, $\mathcal{P^m}$, and $\rho$. Then, in each generation, the parameter values are randomly sampled inside uniform probability distributions between the specified bounds. This approach outperformed the common F-Race-tuned BRKGA in problems with simple decoding strategies.

\section{APIs for BRKGA}
\label{s_api}

To simplify the implementation of BRKGAs, \citet{Toso2015:C++_app_BRKGA} proposed an \textit{Application Programming Interface (API)} in C++ for BRKGA. The API is efficient and user-friendly. The library is portable and automatically handles several aspects of the BRKGA, including population management and evolutionary dynamics. The API is implemented in C++ and uses an object-oriented architecture. In systems with available \textit{OpenMP} \citep{openmp}, the API enables parallel decoding of random key vectors. The user only needs to implement the decoder and specify the stopping criteria, restart and population exchange mechanisms, as well as the parameters of the algorithm. The  API is open source and can be downloaded from \url{http://github.com/rfrancotoso/brkgaAPI}. 

\citet{Andrade2021:BRKGA_MP_IPR} modernized \citet{Toso2015:C++_app_BRKGA} 
leveraging
the BRKGA with multi-parent crossover and implicit path relinking, alongside a multi-population strategy, warm start solution injection, restart, and partial restart features. This framework is also easy to use, as it automatically handles all of those features during the evolutionary process, with the user needing to define only the parameters, decoder, and instance class. The C++ version is available at \url{https://github.com/ceandrade/brkga_mp_ipr_cpp}, the Python version at \url{https://github.com/ceandrade/brkga_mp_ipr_python}, and the Julia package at \url{https://github.com/ceandrade/BrkgaMpIpr.jl}.

One last API, called BRKGACuda 2.0, is presented by \citet{Oliveira2024:brkgacuda20}. This novel application is optimized to work with parallelism and GPUs, significantly accelerating the algorithm. The API provides all of the classical BRKGA evolutionary operators, with the user only needing to specify the decoder. This C++ algorithm can be found at \url{https://github.com/discovery-unicamp/brkga-cuda}.

\section{Conclusion}
\label{s_concl}

This chapter reviewed random-key genetic algorithms, covering both the unbiased and biased variants. After introducing Bean's algorithm \citep{Bean1994:random_keys}, this chapter highlights the minor differences between the RKGA and the BRKGA that contribute to the improved performance of the biased variant. A model for implementing the BRKGA is described, followed by a discussion of optional procedures within the framework. The use of hybrid methods is also detailed, as well as the customization of the algorithm regarding decoding processes and tuning methodologies. Finally, this chapter concludes by presenting three APIs for BRKGA that allow for easy implementation of the algorithm.

\section*{Acknowledgments}

This work was supported by the Brazilian National Council for Scientific and Technological Development (CNPq) under Grant [number 312212/2021-6]; Brazilian Coordination for the Improvement of Higher Level Personnel (CAPES) under Grants [number 001 and 88887.815411/2023]; the Carlos Chagas Filho Research Support Foundation of the State of Rio de Janeiro (FAPERJ) under Grants [numbers 211.086/2019, 211.389/2019, and 211.588/2021].

\bibliographystyle{plainnat}
\bibliography{bibliography,bibliometric,brkga_reviews,new_papers,revBRKGA}

\end{document}